\documentclass[letterpaper, 10 pt, conference]{ieeeconf}  

\IEEEoverridecommandlockouts                              

\usepackage{graphicx}
\usepackage{multirow}
\usepackage{amsmath,amsfonts}
\usepackage{algorithmic}
\usepackage{algorithm}
\usepackage{array}
\usepackage[caption=false,font=normalsize,labelfont=sf,textfont=sf]{subfig}
\usepackage{textcomp}
\usepackage{stfloats}
\usepackage{url}
\usepackage{verbatim}
\usepackage{graphicx}
\usepackage{cite}    
\makeatletter
\let\NAT@parse\undefined
\makeatother
\usepackage{colortbl}
\usepackage{xcolor}
\usepackage{hyperref}
\hypersetup{hypertex=true,
    colorlinks=true,
    linkcolor=red,
    anchorcolor=red,
    citecolor=red}
    
\usepackage{booktabs}
\usepackage{makecell}
\usepackage{booktabs}
\usepackage{bbm}
\usepackage{multirow}
\usepackage{diagbox}

\definecolor{Dark}{rgb}{0.8902,0.9098,0.9412}
\definecolor{correct}{RGB}{173, 173, 173}
\definecolor{incorrect}{RGB}{192, 0, 0}

\usepackage{pifont}
\newcommand{\cmark}{\ding{51}}
\newcommand{\xmark}{\ding{55}}
\usepackage[capitalize]{cleveref}
\makeatother

\title{\LARGE \bf
SplatPose: Geometry-Aware 6-DoF Pose Estimation from Single RGB Image via 3D Gaussian Splatting
}

\author{Linqi Yang, Xiongwei Zhao, Qihao Sun, Ke Wang, Ao Chen, Peng Kang
\thanks{This work is supported in part by the Dreams Foundation of Jianghuai Advance Technology Center (NO.2023-ZM01Z026)}
\thanks{Linqi Yang, Qihao Sun, Ke Wang, Ao Chen are with State Key Laboratory of Robotics and System, Harbin Institute of Technology, Harbin 150006, China (email:23S008025@stu.hit.edu.cn; 23S008047@stu.hit.edu.cn; wangke@hit.edu.cn; 3466509213@qq.com).}%
\thanks{Linqi Yang, Qihao Sun, Ke Wang are also with Zhengzhou Research Institute, Harbin Institute of Technology, Zhengzhou 450000, China.}%
\thanks{Xiongwei Zhao is with School of Information Science and Technology, Harbin Institute of Technology (Shen Zhen), Shenzhen 518071, China (email: xwzhao@stu.hit.edu.cn).}
\thanks{Peng Kang is with the Jianghuai Advance Technology Center, Hefei 230000, China (emal: kangpeng@stu.hit.edu.cn).}
\thanks{Linqi Yang and Xiongwei Zhao contributed equally to this work.}
\thanks{Ke Wang is the corresponding author.}
}

\begin{document}

\maketitle
\thispagestyle{empty}
\pagestyle{empty}

\begin{abstract}
6-DoF pose estimation is a fundamental task in computer vision with wide-ranging applications in augmented reality and robotics. Existing single RGB-based methods often compromise accuracy due to their reliance on initial pose estimates and susceptibility to rotational ambiguity, while approaches requiring depth sensors or multi-view setups incur significant deployment costs. To address these limitations, we introduce SplatPose, a novel framework that synergizes 3D Gaussian Splatting (3DGS) with a dual-branch neural architecture to achieve high-precision pose estimation using only a single RGB image. Central to our approach is the Dual-Attention Ray Scoring Network (DARS-Net), which innovatively decouples positional and angular alignment through geometry-domain attention mechanisms, explicitly modeling directional dependencies to mitigate rotational ambiguity. Additionally, a coarse-to-fine optimization pipeline progressively refines pose estimates by aligning dense 2D features between query images and 3DGS-synthesized views, effectively correcting feature misalignment and depth errors from sparse ray sampling. Experiments on three benchmark datasets demonstrate that SplatPose achieves state-of-the-art 6-DoF pose estimation accuracy in single RGB settings, rivaling approaches that depend on depth or multi-view images.

\end{abstract}

\section{INTRODUCTION}

\PARstart{6}{-DOF} pose estimation, which calculates the 3D position and orientation of a camera in relation to objects or scenes, is fundamental in fields such as robotics and augmented reality.
Although methods based on RGB-D data or point clouds~\cite{zhao2024pnerfloc, he2020pvn3d} have garnered significant attention, their reliance on depth sensors introduces significant limitations, including high hardware costs and sensitivity to challenging material properties (e.g., transparent or reflective surfaces).
In contrast, monocular RGB-based approaches~\cite{tremblay2018deep, lin2022single} offer broader applicability but face inherent trade-offs between scalability, accuracy, and computational efficiency.

A critical challenge in existing RGB-based pose estimation frameworks lies in their dependency on resource-intensive data representations.
For instance, instance-level object pose estimation methods~\cite{sundermeyer2018implicit, peng2019pvnet} require textured 3D models of target objects during training, limiting scalability. 
Neural Radiance Fields (NeRF)~\cite{mildenhall2021nerf} pioneered scene reconstruction through differentiable volume rendering, enabling pose estimation via photometric optimization.
However, NeRF-based methods suffer from two critical limitations: (1) Their implicit volumetric representation requires computationally intensive ray marching, making real-time applications impractical; (2) They demand dense multi-view images for training, which contradicts the single-image inference requirement in most pose estimation scenarios.
Although variants like iNeRF~\cite{yen2021inerf} and Parallel iNeRF~\cite{lin2023parallel} attempt to address these issues, they remain heavily dependent on the initial pose and are susceptible to local minima.

Recent advances in 3D Gaussian Splatting (3DGS)~\cite{yang2024deformable, yu2024mip} have emerged as a paradigm shift, offering explicit scene modeling through anisotropic 3D Gaussian primitives.
Unlike NeRF’s implicit volumetric approach, 3DGS supports real-time rendering by utilizing rasterization and retains photorealistic visual fidelity, making it particularly attractive for pose estimation tasks.
However, methods like SplatLoc~\cite{zhai2024splatloc} and 3DGS-ReLoc~\cite{jiang20243dgs} rely on dense depth data and multi-view images to reconstruct scenes and retrieve the initial pose, resulting in substantial storage and data collection costs.
Conversely, single RGB image-based methods, such as 6DGS~\cite{bortolon20246dgs}, eliminate the need for depth sensors or multi-frame databases by directly leveraging 3DGS’s differentiable rendering through rendering inversion.
However, 6DGS’s Gaussian ellipsoid-based ray sampling strategy introduces rotational ambiguity due to its bias toward rays with minimal perpendicular distance to the camera's optical center, while neglecting angular offsets. 
These shortcomings highlight a fundamental trade-off: single RGB-based methods rely on initial pose estimation and introduce rotational ambiguity, while methods that incorporate depth or multiple views incur prohibitive storage and data acquisition costs.

\begin{figure*}[t]
    \centering
    \includegraphics[width=1.0\hsize]{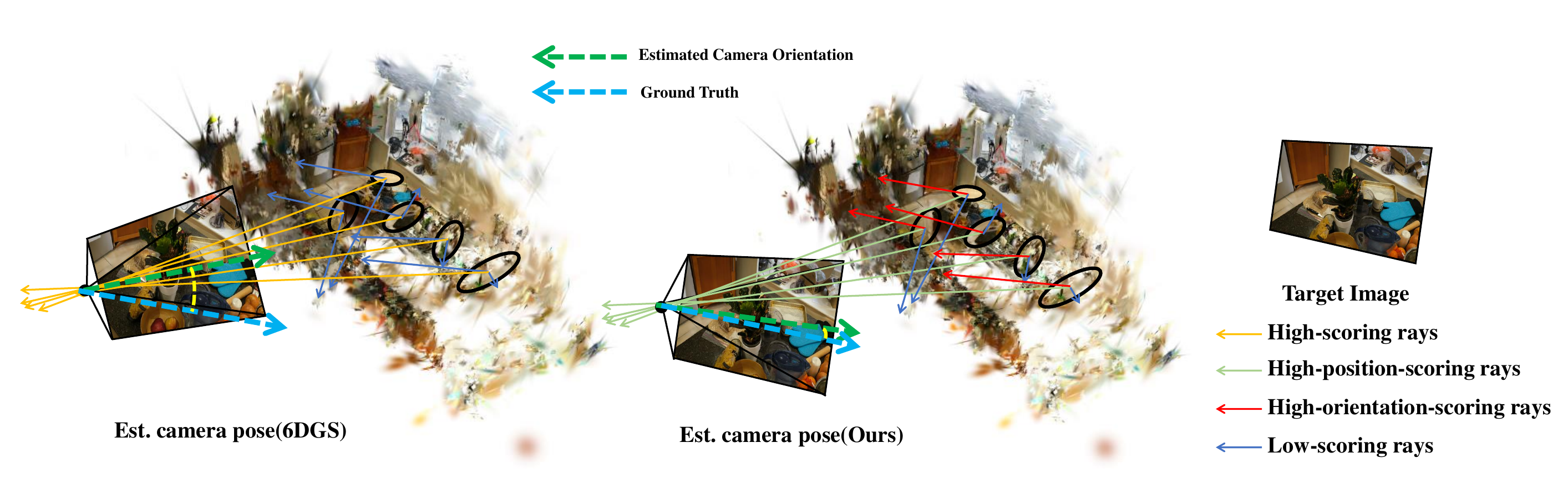}
    \caption{Comparison of pose estimation between 6DGS~\cite{bortolon20246dgs} and SplatPose. 6DGS selects high-scoring rays solely based on proximity to the camera's optical center, while SplatPose, via DARS-Net, refines pose estimation by incorporating both high-position-scoring rays (closer to the optical center) and high-orientation-scoring rays (aligned with the camera orientation), ultimately achieving smaller rotational errors compared to 6DGS.}
    \label{intro}
\end{figure*}

To address these challenges, we introduce SplatPose, a novel framework aimed at solving the problem of 6-DoF pose estimation using a single RGB image.
First, SplatPose proposes the Dual-Attention Ray Scoring Network (DARS-Net), which introduces a refined geometry scoring mechanism by decomposing the ray score into two independent components: position score and orientation score.
By leveraging high-position-scoring rays and high-orientation-scoring rays to independently determine the camera's position and orientation, DARS-Net effectively overcomes the rotational ambiguity, achieving significant improvements in both translational and rotational accuracy, as illustrated in~\cref{intro}.
Second, SplatPose designs an innovative 6-DoF pose estimation pipeline within a coarse-to-fine framework, which employs an effective feature point matching technique to refine the coarse pose initially derived from 3DGS rays.
It represents a robust and scalable solution that pushes the boundaries of pose estimation based on 3D models.
In summary, the key contributions of our proposed method are outlined as follows:
\begin{itemize}
    
    \item We propose the Dual-Attention Ray Scoring Network (DARS-Net), which leverages attention mechanisms to decompose ray scoring into position and orientation components, effectively mitigating rotational ambiguity in 6-DoF pose estimation from a single RGB image.
    \item We introduce a novel coarse-to-fine pipeline that employs an efficient keypoint matching technique to refine the coarse pose estimated from 3DGS rays.
    \item The proposed SplatPose achieves state-of-the-art 6-DoF pose estimation results on three public Novel View Synthesis benchmarks, outperforming existing single RGB-based pose estimation methods while achieving performance comparable to depth- and multi-view-based approaches.

\end{itemize}

\section{RELATED WORKS}


\subsection{Pose estimation based on Neural Radiance Fields}
iNeRF~\cite{yen2021inerf} presented a framework employing NeRF to estimate 6-DoF poses by matching rendered images to target images through minimizing photometric errors.
However, it tends to become trapped in local minima, leading to advancements like Parallel iNeRF~\cite{lin2023parallel}, which optimizes multiple candidate poses in parallel. 
NeMo+VoGE~\cite{wang2021nemo, wang2022voge} uses volumetric Gaussian reconstruction kernels but relies on ray marching and iterative optimization, requiring training on multiple objects. 
In comparison, our approach utilizes a single-object 3DGS model, making it more efficient.
CROSSFIRE~\cite{moreau2023crossfire} incorporates learned local features to mitigate local minima but still relies on accurate initial pose priors.
IFFNeRF~\cite{bortolon2024iffnerf} proposes NeRF model inversion to re-render images matching a target view but overlooks unique 3DGS characteristics, such as ellipsoid elongation, rotation, and non-uniform spatial distribution, which our approach effectively addresses.

\subsection{Pose estimation based on 3D Gaussian Splatting}
3DGS-ReLoc~\cite{jiang20243dgs} pioneers LiDAR-camera fused 3DGS mapping using KD-trees and 2D voxel grids, employing NCC for coarse alignment and PnP for pose refinement.
GSLoc~\cite{niu2024hgsloc} tackles photometric loss non-convexity via coarse-to-fine optimization, backpropagating gradients through 3DGS to refine sparse feature-based initializations.
Meanwhile, SplatLoc~\cite{zhai2024splatloc} proposes a hybrid framework merging explicit 3DGS maps with learned descriptors, using saliency-driven landmark selection and anisotropic regularization to ensure accurate 2D-3D matching with compactness. 
These methods predominantly rely on depth information for 3D Gaussian scene reconstruction or necessitate multi-view image sequences. 
In contrast, 6DGS~\cite{bortolon20246dgs} eliminates dependency on pose initialization by inverting the 3DGS rendering process, thereby achieving single-RGB-image-based 6-DoF camera pose estimation. 
However, its Gaussian ellipsoid-based ray sampling strategy introduces rotational ambiguity, a limitation fundamentally addressed in our approach by the Dual-Attention Ray Scoring Network (DARS-Net), which resolves this geometric ambiguity through a meticulously designed dual-branch attention mechanism.

\subsection{Correspondence matching}
Conventional methods for 6-DoF image matching predominantly rely on feature-based approaches, including both classical hand-crafted descriptors such as SIFT~\cite{lowe1999object} and modern deep learning techniques like SuperGlue~\cite{sarlin2020superglue} and TransforMatcher~\cite{kim2022transformatcher}. 
SuperGlue utilizes a Graph Neural Network (GNN) to enhance feature focus and applies the Sinkhorn algorithm~\cite{cuturi2013sinkhorn} for establishing correspondences. 
TransforMatcher\cite{kim2022transformatcher} incorporates global match-to-match attention, facilitating accurate localization of correspondences.
Additionally, feature equivariance techniques~\cite{lee2022self,lee2023learning} have been developed to enhance robustness by ensuring features remain consistent under transformations. 
These approaches, however, presume that the two feature ensembles intrinsically reside within a homogeneous data modality, usually derived directly from image data. 
In 3DGS-based approaches, the matching challenge differs, as it entails associating pixels with rays originating from the Ellicells. 
While OnePose++~\cite{he2022onepose++} employs point cloud-image matching and CamNet~\cite{ding2019camnet} directly regresses poses, both require extensive multi-scene training ($\geq$500 images). 
To overcome these limitations, we utilize the proposed attention model to efficiently manage associations between rays and pixels, and achieve greater data efficiency by operating with significantly fewer images (approximately 100 or less) used exclusively during training.

\begin{figure*}[t]
    \centering
    \includegraphics[width=1.0\hsize]{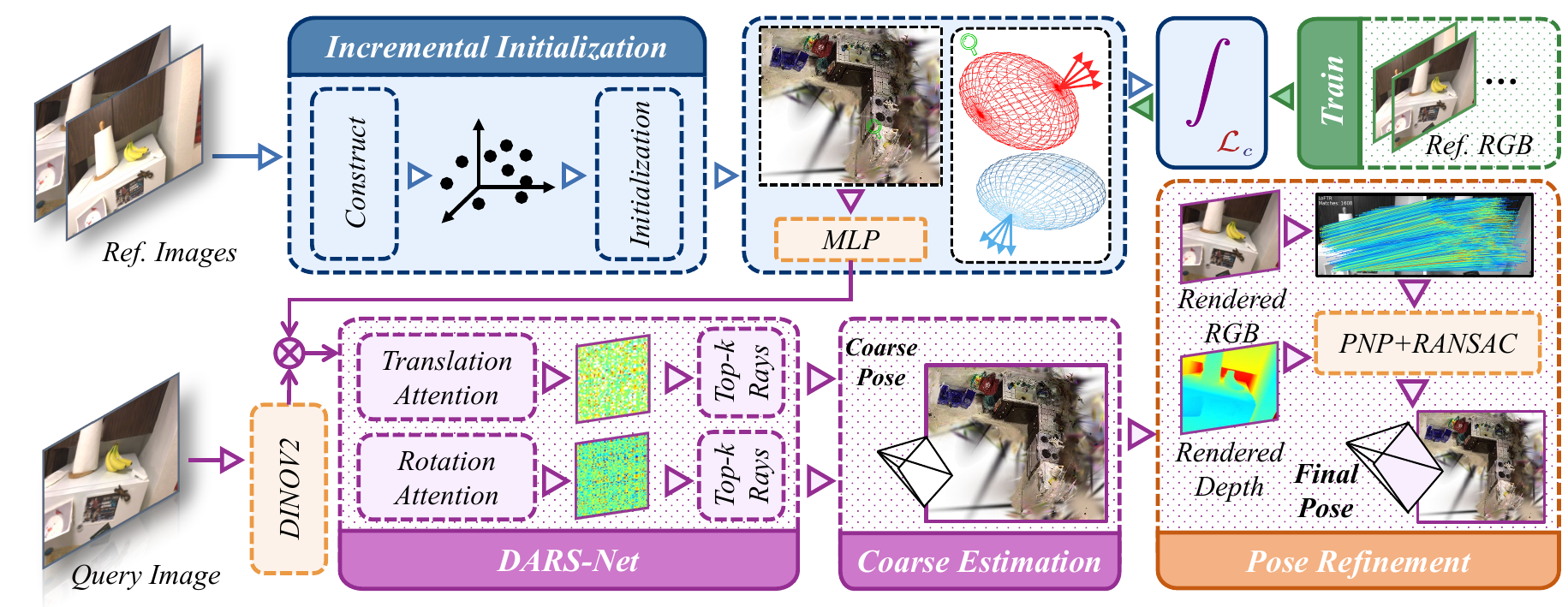}
    \caption{An overview of our SplatPose pipeline. Our framework is composed of three key stages: (1) 3D Gaussian Scene Representation, where a 3DGS scene map is constructed from sparse point clouds to initialize the scene representation; (2) DARS-Net and Coarse Estimation, which decouples ray scoring into translation and rotation attention mechanisms, independently computing position and orientation scores for cast rays, selecting top-k rays based on these scores, and leveraging them to estimate the camera's position and orientation; and (3) Pose Refinement, where a synthetic scene view is rendered using the coarse pose, and keypoints are matched between the rendered view and the query image to refine the camera pose.}
    \label{pipeline}
\end{figure*}

\section{METHODOLODY}
In this part, we present the pipelines for reconstruction and pose estimation in our method, as illustrated in~\cref{pipeline}. 
In~\cref{A}, we begin by presenting the 3D Gaussian scene representation. 
Then, we introduce the Dual-Attention Ray Scoring Network(~\cref{B}). 
Finally, the coarse pose estimation and pose refinement are shown in~\cref{C} and~\cref{D}.

\subsection{3D Gaussian Scene Representation}\label{A}
3D Gaussian Splatting represents a scene explicitly by employing a set of anisotropic 3D Gaussian primitives. 
Each Gaussian primitive is characterized in the world space by a mean vector $\mu \in \mathbb{R}^3$ and a covariance matrix $\Sigma \in \mathbb{R}^{3\times3}$, as described by:
\begin{equation}
    G(\mu, \Sigma) = e^{-\frac{1}{2}(x - \mu)^T \Sigma (x - \mu)}.
\end{equation}
To ensure the covariance matrix $\Sigma$ retains its physical validity during optimization, it is expressed as the decomposition of a scaling matrix \textit{S} and a rotation matrix \textit{R}, as proposed in~\cite{kerbl20233d}:
\begin{equation}
    \Sigma = R S S^T R^T,
\end{equation}
where the scaling matrix \textit{S} is derived from a 3D scale vector \textbf{s}, \textit{S} = diag([\textbf{s}]), and the rotation matrix \textit{R} is parameterized using a quaternion.

Following the method in~\cite{zwicker2001ewa}, the 3D Gaussians are projected into the 2D image plane for rendering. 
The covariance matrix in the camera coordinate system is computed utilizing the viewing matrix \textit{W} alongside the Jacobian \textit{J} derived from the affine approximation of the projective transformation, as follows:
\begin{equation}
    \widetilde{\Sigma} = J W \Sigma W^T J^T.
\end{equation}
The corresponding 2D Gaussian distribution $\hat{G}(\widetilde{\mu}, \widetilde{\Sigma})$ is then derived from the 2D pixel location $\widetilde{\mu}$ of the 3D Gaussian center and the projected covariance matrix $\widetilde{\Sigma}$.

For novel view synthesis and fast rasterization-based rendering, each 3D Gaussian primitive is associated with an opacity value $\sigma \in \mathbb{R}$ and a color $\textit{c} \in \mathbb{R}^3$, represented using spherical harmonics (SH) coefficients. 
To achieve photorealistic rendering, the differentiable rasterizer employs alpha blending~\cite{max1995optical}, which accumulates Gaussian properties and opacity values $\sigma$ for each pixel by traversing the ordered primitives. 
Specifically, the color properties are computed as follows:
\begin{equation}
    \hat{I} = \sum_{i=1}^{N} c_{i} \cdot \alpha_{i} \cdot \prod_{j=1}^{i-1} (1 - \alpha_{j}),
\end{equation}
where $\hat{I}$ is the rendered color. Here, $\alpha_{i}=\hat{G}(\widetilde{\mu}, \widetilde{\Sigma})$ represents the opacity contribution of the $i$-th Gaussian to the pixel, $\prod_{j=1}^{i-1} (1 - \alpha_{j})$ denotes the accumulated transmittance, and \textit{N} is the total number of Gaussian primitives contributing to the pixel during the splatting process.
\subsection{Dual-Attention Ray Scoring Network}\label{B}
The primary challenge of rotational ambiguity in monocular 6-DoF pose estimation arises from the undifferentiated treatment of spatial and angular information in conventional ray scoring. 
To address this, we propose the Dual-Attention Ray Scoring Network (DARS-Net), which independently estimates camera position and orientation by leveraging dual attention mechanisms to model position and orientation scores for cast rays. 
Specifically, we generate multiple cast rays \textbf{r} for each Gaussian ellipsoid and determine a subset of \textbf{r} that corresponds to the target image $\textbf{I}_t$. 
Two attention maps, $\textit{A}_p$ and $\textit{A}_o$, compute position scores $\hat{s}_p$ and orientation scores $\hat{s}_o$ by evaluating the correlation between rays and image pixels. 
The top \textit{K} rays from $\textit{A}_p$ estimate the camera position, while the top \textit{K} rays from $\textit{A}_o$ determine the orientation.

Ray features $\textbf{R} \in \mathbb{R}^{N \times C}$, where $\textit{C}$ indicates the feature dimension, while $\textit{N}$ represents the total count of rays, are extracted using an augmented Multi-Layer Perceptron (MLP) architecture incorporating spatial coordinate embedding \cite{tancik2020fourier}, boosting the network’s ability to differentiate features. 
Image features are extracted from $\textbf{I}_t$ using the pre-trained DINOv2\cite{oquab2023dinov2} backbone, producing feature sets $\textbf{F} \in \mathbb{R}^{M \times C}$, where $\textit{M} = \textit{W} \times \textit{H}$, \textit{W} represents the image width, and \textit{H} represents the image height. 
These are processed through attention modules $\textit{A}_p(\textbf{R}, \textbf{F}) \in \mathbb{R}^{M \times N}$ and $\textit{A}_o(\textbf{R}, \textbf{F}) \in \mathbb{R}^{M \times N}$, where ray features function as queries, while image features operate as keys. 
The resulting attention maps are optimized by performing row-wise summation and transforming them into per-ray correlation scores, respectively defined as position scores $\hat{s}_p=\sum_{i=1}^{M}\textit{A}_{pi}$ and orientation scores $\hat{s}_o=\sum_{i=1}^{M}\textit{A}_{oi}$. 
During inference, the top \textit{K} rays with the highest position scores predict the camera position, while those with the highest orientation scores determine the orientation.

The predicted scores $\hat{s}_p$ and $\hat{s}_o$ are supervised employing identical images from the 3DGS training set, under supervision based on the distance from the camera origin to its projection on the corresponding ray, along with the angle between the camera's orientation and the ray's direction. 
The projection is computed as $L = \max((\textbf{P} - \textbf{r}_o)\textbf{r}_d, 0)$, where $\textbf{P}$ is the camera position, $\textbf{r}_o$ the ray origin, and $\textbf{r}_d$ the ray direction. 
The distance is given by $d = |(\textbf{r}_o + L\textbf{r}_d) - \textbf{P}|_2$, with $d = 0$ indicating that the ray intersects the optical center. 
The angle is calculated as $\theta = \arccos\left(\frac{-\textbf{Q} \cdot \textbf{r}_d}{|\textbf{Q}| \cdot |\textbf{r}_d|}\right)$, where $\textbf{Q}$ is the camera orientation. 
These distances and angles are mapped to attention map scores for supervision.
We map these distances and angels to attention map scores using:
\begin{equation}
    \begin{aligned}
    \alpha = 1 - \tanh\left(\frac{d}{\gamma}\right); \quad s_p = \alpha \frac{M}{\sum \alpha};\\
    \beta = 1 - \tanh\left(\frac{\theta}{\gamma}\right); \quad s_o = \beta \frac{M}{\sum \beta}.
    \end{aligned}
\end{equation}
Here, $\gamma$ regulates the allocation of rays to a given camera. 
Additionally, to compute the attention maps, the ground truth scores must be normalized due to the softmax operation. 
To optimize the predicted position scores $\hat{s}_p$ and orientation scores $\hat{s}_o$ against the computed ground truth position scores $s_p$ and orientation scores $s_o$, we employ the $L_2$ loss, formulated as:
\begin{equation}
    \begin{aligned}
        \mathcal{L}_p &= \frac{1}{N} \sum_{i=1}^{N} \| \hat{\textit{s}}_{p_{i}} - \textit{s}_{p_{i}} \|_2, \\
        \mathcal{L}_o &= \frac{1}{N} \sum_{i=1}^{N} \| \hat{\textit{s}}_{o_{i}} - \textit{s}_{o_{i}} \|_2, \\
        \mathcal{L} &= \mathcal{L}_p + \mathcal{L}_o.
    \end{aligned}
\end{equation}
Where $\mathcal{L}_p$ represents the positional score loss, while $\mathcal{L}_o$ denotes the orientation score loss. In every training iteration, a predicted image and pose are utilized to estimate the 3DGS model.

\subsection{Coarse Pose Estimation}\label{C}
The predicted position scores $\hat{s}_p$ and orientation scores $\hat{s}_o$ are independently used to determine the top $K$ most relevant rays for position and direction respectively, with the selection restricted to a maximum of one ray per ellipsoid.
The camera position is computed at the intersection of the selected rays and formulated as a weighted least-squares optimization problem. 
Due to discretization noise from the DARS-Net, 3D rays rarely converge at a single point. 
Therefore, the problem is addressed by minimizing the summation of squared normal distances. 
For each selected ray $\textbf{r}_i$ with $i=1...K$, the error is defined as the squared distance between the predicted camera position $\hat{\textbf{P}}$ and its orthogonal projection onto $\textbf{r}_i$:
\begin{equation}\label{1}
    \sum_{i=1}^{K} \left( (\hat{\textbf{P}} - \textbf{r}_{o,i})^T (\hat{\textbf{P}} - \textbf{r}_{o,i}) - ((\hat{\textbf{P}} - \textbf{r}_{o,i})^T \textbf{r}_{d,i})^2 \right),
\end{equation}
where $\textbf{r}_{o,i}$ represents the origin of the $i$-th ray, and $\textbf{r}_{d,i}$ denotes its corresponding direction. To minimize \cref{1}, the equation is differentiated with respect to $\hat{\textbf{P}}$, yielding:
\begin{equation}
    \hat{\textbf{P}} = \sum_{i=1}^{N} \hat{\textit{s}}_{p,i} (\mathbb{I} - \textbf{r}_{d,i} \textbf{r}_{d,i}^T) \textbf{r}_{o,i},
\end{equation}
where $\mathbb{I}$ denotes the identity matrix, and $\hat{\textit{s}}_{p,i}$ represent the predicted position scores. 
This formulation can be resolved as a weighted linear system.

The camera orientation is computed as the negative weighted sum of the direction vectors of the selected rays, with the weights determined by their predicted orientation scores $\hat{\textit{s}}_o$. 
The resulting orientation vector $\hat{\textbf{Q}}$ is expressed as:
\begin{equation}
    \hat{\textbf{Q}} = -\frac{\sum_{i=1}^{N} \hat{\textit{s}}_{o,i} \textbf{r}_{d,i}}{\left\| \sum_{i=1}^{N} \hat{\textit{s}}_{o,i} \textbf{r}_{d,i} \right\|},
\end{equation}
where $\hat{\textit{s}}_{o,i}$ denotes predicted orientation scores. The normalization ensures that the computed orientation vector has unit magnitude.

\subsection{Pose Refinement}\label{D}
The coarse pose estimation process relies on ray sampling; however, the presence of noisy rays often prevents even high-scoring rays from precisely traversing the optical center of the camera, leading to inaccuracies in position and orientation estimation. 
This limitation imposes an inherent upper bound on the accuracy of pose estimation, necessitating further refinement. 
The refinement process begins with the extraction and matching of 2D feature points using LoFTR~\cite{sun2021loftr}, a transformer-based feature matching method. 
Note that LoFTR here can be replaced with any other feature matcher. 
Using the coarse pose estimate, the 3D Gaussian primitives are mapped onto the 2D image plane to generate a synthetic rendering of the scene. 
LoFTR then computes high-quality 2D-2D correspondences between the query image and the rendered view, resulting in a set of matched keypoints. 
These 2D-2D correspondences are leveraged to compute 2D-3D correspondences by back-projecting the 2D keypoints in the rendered view to their corresponding 3D coordinates using the depth information of the 3D Gaussian primitives and the camera intrinsics. 
The resulting 2D-3D correspondences are then used to estimate the refined camera pose through a Perspective-n-Point (PnP) algorithm.

\vspace{1cm}
\begin{table*}[t]\Large
    \caption{The \textbf{Mean Angular Error (MAE)} and \textbf{Mean Translation Error (MTE)} for 6-DoF pose estimation are evaluated on \textit{Mip-NeRF $360^\circ$} in degrees and units $u$, where $1u$ is the object's largest dimension. Lower values indicate better performance. \textbf{[up]}: Fixed pose prior (from~\cite{yen2021inerf}). \textbf{[middle]}: Random pose prior. \textbf{[bottom]}: No pose prior. {\color{red}{Red}}: Best. {\color{blue}{Blue}}: Second best.}
    \label{tab:mipnerf360}
    \centering
    \renewcommand\arraystretch{1.2}
    \linespread{1.20}\selectfont
    \resizebox{1.0\textwidth}{!}{
        \begin{tabular}{r|c|ccccccc}
            \toprule
            \textbf{Method} & \textbf{Avg~$\downarrow$} & \textbf{Bicycle} & \textbf{Bonsai} & \textbf{Counter} & \textbf{Garden} & \textbf{Kitchen} & \textbf{Room} & \textbf{Stump} \\
            \midrule\midrule
            iNeRF~\cite{yen2021inerf} & $37.3 / 0.172$ & $39.5 / 0.116$ & $51.3 / 0.228$ & $40.7 / 0.324$ & $31.0 / 0.121$ & $38.2 / 0.113$ & $38.8 / 0.274$ & $21.4 / 0.030$ \\
            NeMo + VoGE~\cite{wang2022voge} & $40.9 / 0.036$ & $43.8 / 0.015$ & $52.5 / 0.036$ & $45.6 / 0.072$ & $31.8 / 0.026$ & $41.6 / 0.042$ & $44.9 / 0.045$ & $26.3 / 0.016$ \\
            Parallel iNeRF~\cite{lin2023parallel} & $28.9 / 0.146$ & $35.9 / 0.116$ & $41.1 / 0.223$ & $24.7 / 0.212$ & $18.2 / 0.090$ & $37.3 / 0.109$ & $30.7 / 0.257$ & $14.8 / 0.016$ \\
            \midrule
            iNeRF~\cite{yen2021inerf} & $85.0 / 0.292$ & $76.6 / 0.217$ & $96.7 / 0.385$ & $70.3 / 0.487$ & $72.8 / 0.210$ & $100.2 / 0.266$ & $91.6 / 0.444$ & $86.9 / 0.035$ \\
            NeMo + VoGE~\cite{wang2022voge} & $103.8 / 0.058$ & $111.8 / 0.038$ & $98.9 / 0.073$ & $98.1 / 0.139$ & $89.2 / 0.038$ & $122.2 / 0.082$ & $110.0 / \color{red}{0.010}$ & $96.3 / 0.025$ \\
            Parallel iNeRF~\cite{lin2023parallel} & $58.0 / 0.218$ & $44.4 / 0.150$ & $58.2 / 0.298$ & $42.1 / 0.435$ & $60.0 / 0.144$ & $65.0 / 0.193$ & $63.5 / 0.271$ & $72.6 / 0.033$ \\
            \midrule
            6DGS~\cite{bortolon20246dgs} & $24.3 / 0.022$ & $12.1 / 0.010$ & $10.5 / 0.038$ & $19.6 / 0.043$ & $37.8 / 0.015$ & $23.2 / 0.018$ & $38.3 / 0.019$ & $28.3 / 0.009$ \\
            Ours(Only DARS-Net) & $\color{blue}{11.1 / 0.012}$ & $\color{blue}{9.14 / 0.010}$ & $\color{blue}{5.79 / 0.020}$ & $\color{blue}{9.65 / 0.022}$ & $\color{blue}{21.9 / 0.008}$ & $\color{blue}{7.91 / 0.009}$ & $\color{blue}{8.79 / 0.013}$ & $\color{blue}{14.3 / 0.005}$ \\
            Ours(DARS-Net + Pose Refinement) & $\color{red}{1.06 / 0.007}$ & $\color{red}{0.17 / 0.003}$ & $\color{red}{0.73 / 0.006}$ & $\color{red}{0.52 / 0.015}$ & $\color{red}{2.55 / 0.005}$ & $\color{red}{0.47 / 0.008}$ & $\color{red}{2.44 / 0.010}$ & $\color{red}{0.53 / 0.002}$ \\
            \bottomrule
        \end{tabular}
    }
\end{table*}

\begin{table*}[t]\scriptsize
    \caption{The \textbf{Mean Angular Error (MAE)} and \textbf{Mean Translation Error (MTE)} for 6-DoF pose estimation are evaluated on \textit{Tanks$\&$Temples} in degrees and units $u$, where $1u$ is the object's largest dimension. Lower values indicate better performance. \textbf{[up]}: Fixed pose prior (from~\cite{yen2021inerf}). \textbf{[middle]}: Random pose prior. \textbf{[bottom]}: No pose prior. {\color{red}{Red}}: Best. {\color{blue}{Blue}}: Second best.}
    \label{tab:tanks}
    \centering
    \renewcommand\arraystretch{1.2}
    \resizebox{0.95\textwidth}{!}{
        \begin{tabular}{r|c|ccccccc}
            \toprule
            \textbf{Method} & \textbf{Avg~$\downarrow$} & \textbf{Barn} & \textbf{Caterpillar} & \textbf{Family} & \textbf{Ignatius} & \textbf{Truck} \\
            \midrule\midrule
            iNeRF~\cite{yen2021inerf} & $35.0 / 0.452$ & $26.5 / 0.208$ & $42.9 / 0.166$ & $42.8 / 0.794$ & $31.4 / 0.723$ & $31.6 / 0.370$ \\
            NeMo + VoGE~\cite{wang2022voge} & $53.6 / 0.965$ & $51.2 / 0.752$ & $52.6 / 0.516$ & $58.4 / 1.130$ & $51.2 / 1.193$ & $54.6 / 1.236$ \\
            Parallel iNeRF~\cite{lin2023parallel} & $24.7 / 0.346$ & $22.9 / 0.131$ & $25.2 / 0.138$ & $22.9 / 0.507$ & $23.4 / 0.604$ & $29.4 / 0.351$ \\
            \midrule
            iNeRF~\cite{yen2021inerf} & $90.2 / 1.455$ & $89.2 / 0.682$ & $89.3 / 2.559$ & $93.9 / 1.505$ & $84.1 / 1.489$ & $94.4 / 1.042$ \\
            NeMo + VoGE~\cite{wang2022voge} & $92.6 / 1.457$ & $92.5 / 0.684$ & $90.5 / 2.559$ & $97.0 / 1.506$ & $85.4 / 1.491$ & $97.7 / 1.045$ \\
            Parallel iNeRF~\cite{lin2023parallel} & $91.1 / 1.130$ & $85.2 / 0.572$ & $86.8 / 0.843$ & $99.0 / 2.028$ & $86.9 / 1.326$ & $97.6 / 0.883$ \\
            \midrule
            6DGS~\cite{bortolon20246dgs} & $21.7 / 0.268$ & $30.3 / 0.162$ & $14.5 / 0.027$ & $20.6 / 0.468$ & $15.5 / 0.441$ & $27.5 / 0.242$ \\
            Ours(Only DARS-Net) & $\color{blue}{5.36 / 0.257}$ & $\color{blue}{5.13 / 0.147}$ & $\color{blue}{4.91 / 0.025}$ & $\color{blue}{4.52 / 0.460}$ & $\color{blue}{5.90 / 0.412}$ & $\color{blue}{6.35 / 0.239}$ \\
            Ours(DARS-Net + Pose Refinement) & $\color{red}{2.97 / 0.211}$ & $\color{red}{3.86 / 0.122}$ & $\color{red}{2.00 / 0.023}$ & $\color{red}{3.16 / 0.413}$ & $\color{red}{1.92 / 0.273}$ & $\color{red}{3.90 / 0.227}$ \\
            \bottomrule
        \end{tabular}
    }
\end{table*}

\vspace{1cm}
\begin{table*}[t]\scriptsize
    \caption{The \textbf{Median Angular Error and Median Translation Error} (MAE, MTE) for 6-DoF pose estimation are evaluated on \textit{12Scenes} in degrees and $cm$. Lower values indicate better performance. {\color{red}{Red}}: Best. {\color{blue}{Blue}}: Second best.}
    \label{tab:12scenes}
    \centering
    \renewcommand\arraystretch{1.2}
    \resizebox{0.95\textwidth}{!}{
        \begin{tabular}{r|ccccccccccc}
            \toprule
            & \multicolumn{2}{c}{Apartment 1} & \multicolumn{3}{c}{Apartment 2} & \multicolumn{4}{c}{Office 1}           & \multicolumn{2}{c}{Office 2} \\
            \cmidrule(lr){2-3} \cmidrule(lr){4-6} \cmidrule(lr){7-10} \cmidrule(lr){11-12}
            \textbf{Method} & \textbf{kitchen}         & \textbf{living}      & \textbf{kitchen}  & \textbf{living} & \textbf{luke} & \textbf{gates362} & \textbf{gates381} & \textbf{lounge} & \textbf{manolis} & \textbf{5a}            & \textbf{5b}           \\
            \midrule\midrule
            SCRNet~\cite{li2020hscnet} & $\color{blue}1.3 / 2.3$ & $\color{blue}0.8 / 2.4$ & $\color{blue}1.0 / 2.1$ & $1.8 / 4.2$ & $1.4 / 4.4$ & $0.8 / 2.6$ & $1.4 / {\color{blue}3.4}$ & $\color{blue}0.9 / 2.7$ & $1.0 / {\color{blue}1.8}$ & $\color{blue}1.5 / 3.6$ & ${\color{blue}1.2} / 3.4$\\
            SplatLoc~\cite{zhai2024splatloc} & $\color{red}0.4 / 0.8$ & $\color{red}0.4 / 1.1$ & $\color{red}0.5 / 1.0$ & $\color{red}0.5 / 1.2$ & $\color{red}0.6 / 1.5$ & ${\color{blue}0.5} / {\color{red}1.1}$ & $\color{red}0.5 / 1.2$ & $\color{red}0.5 / 1.6$ & $\color{red}0.5 / 1.1$ & $\color{red}0.6 / 1.4$ & $\color{red}0.5 / 1.5$\\
            Ours & ${\color{red}{0.4}} / 5.3$ & ${\color{red}0.4} / 3.6$ & ${\color{red}0.5} / 2.9$ & ${\color{red}0.5} / {\color{blue}2.8}$ & $\color{blue}0.7 / 4.2$ & ${\color{red}0.3} / {\color{blue}1.9}$ & ${\color{blue}0.7} / 4.0$ & ${\color{red}0.5} / 5.2$ & ${\color{blue}0.6} / 3.1$ & ${\color{red}0.6} / 5.0$ & ${\color{red}0.5} / {\color{blue}3.0}$ \\
            \bottomrule
        \end{tabular}
    }
\end{table*}

\begin{table}[t]\scriptsize
    \caption{Memory usage, training time and inference time of different
        methods on scene \textbf{manolis} from 12-Scenes Dataset.}
    \label{tab:time}
    \centering
    \renewcommand\arraystretch{1.5}
    \resizebox{0.45\textwidth}{!}{
        \begin{tabular}{r|c|c|c}
            \toprule
            \textbf{Method} & \textbf{Memory~$\downarrow$} & \textbf{Training time~$\downarrow$} & \textbf{Inference time~$\downarrow$}\\
            \midrule\midrule
            SCRNet~\cite{li2020hscnet} & $165MB$ & $2days$ & $1min$\\
            SplatLoc~\cite{zhai2024splatloc} & $737MB$ & $25mins$ & $9mins13s$\\
            \rowcolor{gray!20} Ours & $264MB$ & $45mins$ & $6mins20s$\\
            \bottomrule
        \end{tabular}
    }
\end{table}

\begin{figure*}[htbp]
    \centering
    \includegraphics[width=0.9\hsize]{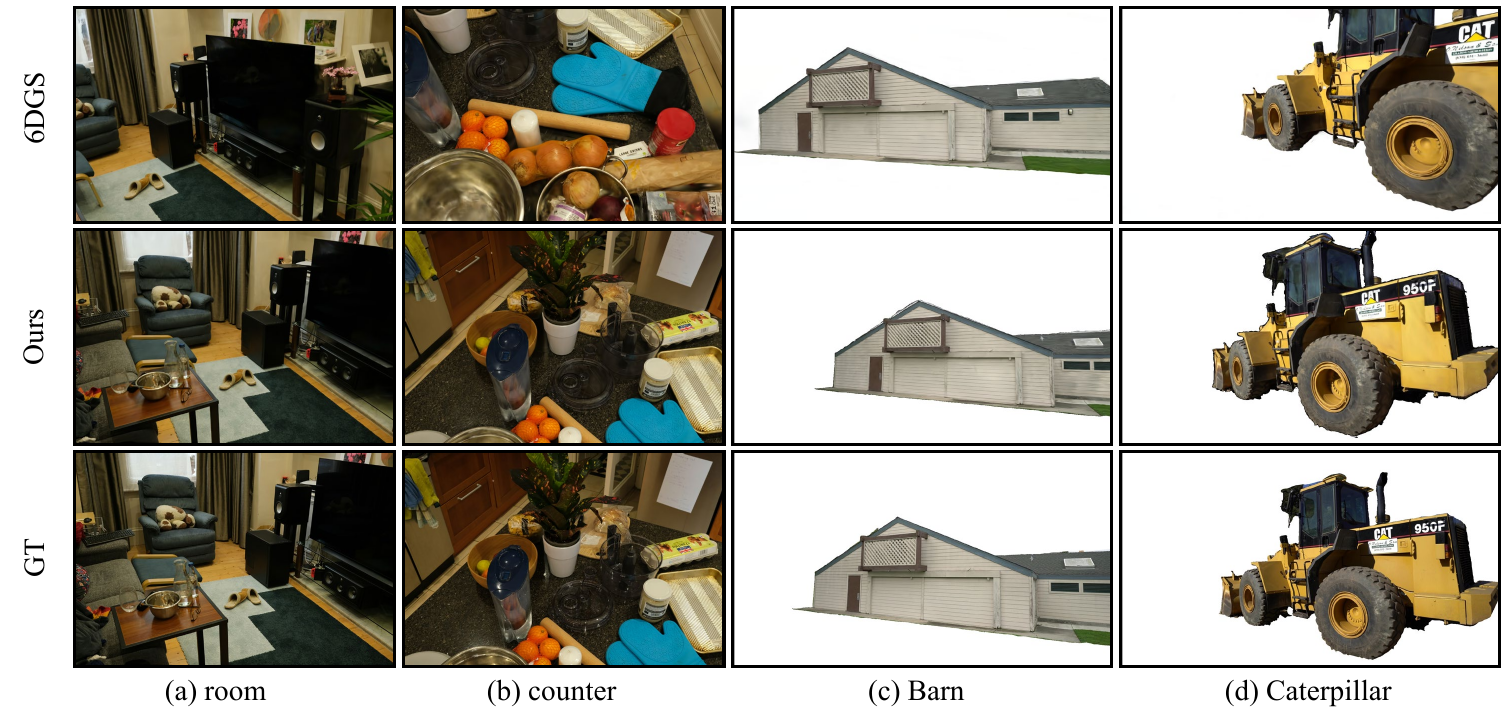}
    \caption{The illustration presents qualitative results from the Mip-NeRF 360° dataset ((a) and (b)) and the Tanks \& Temples dataset ((c) and (d)). From top to bottom, there are results of 6DGS~\cite{bortolon20246dgs}, ours, and ground truth. For each scene, the images are rendered based on the estimated camera poses utilizing the provided 3DGS model.}
    \label{fig:result_compare}
\end{figure*}

\section{EXPERIMENTS}

\subsection{Experimental Setup}
We compare SplatPose with the 3DGS-based method 6DGS~\cite{bortolon20246dgs} and Nerf-based approaches for 6-DoF pose estimation with single RGB image, including iNeRF~\cite{yen2021inerf}, Parallel iNeRF~\cite{lin2023parallel}, and NeMo+VoGE~\cite{wang2022voge}. 
Following the evaluation protocol in~\cite{bortolon20246dgs}, experiments are conducted on Mip-NeRF $360^\circ$~\cite{barron2022mipnerf360} and Tanks$\&$Temples~\cite{knapitsch2017tanks} datasets using their predefined training-test splits. 
We test under two pose initialization scenarios: (i) iNeRF initialization, with uniformly sampled errors in $\left[-40^\circ, +40^\circ\right]$ for rotation and $\left[-0.1, +0.1\right]$ for translation; and (ii) a realistic initialization, where the starting pose is randomly chosen from those used to create the 3DGS model. The second setting evaluates methods under more practical conditions. 
Ablation studies are also performed to validate each system component. 
Pose estimation performance is measured using mean angular error (MAE) and mean translational error (MTE) (see \cref{tab:mipnerf360} and~\cref{tab:tanks}). 

Additionally, to verify SplatPose's robustness in intricate physical environments, we follow the experimental setup of~\cite{zhai2024splatloc}, selecting two Depth- and Multi-View-Based approaches for comparative analysis, including scene coordinate regression approach SCRNet~\cite{li2020hscnet} and recent 3DGS-based visual localization approach SplatLoc~\cite{zhai2024splatloc}, using the 12Scenes dataset\cite{valentin2016learning}.

\textbf{Implementation Details.} The SplatPose framework is implemented using PyTorch, with the attention map trained for 1,500 iterations (approximately 45 minutes) on an NVIDIA GeForce RTX 3090 GPU. 
This optimization employs the Adafactor algorithm~\cite{shazeer2018adafactor}, with a weight decay coefficient of $10^{-3}$. 
To accelerate the training process, 2,000 3DGS ellipsoids are uniformly sampled at each iteration.

\subsection{Datasets}

\textbf{Mip-NeRF} $\mathbf{360^\circ}$\cite{barron2022mipnerf360} includes seven scenes (two outdoor, five indoor) with structured settings and consistent backgrounds. We use the original 1:8 train-test splits from\cite{barron2022mipnerf360}. Following~\cite{lin2023parallel}, all objects are scaled to a unit box, and translation errors are normalized by object size.

\textbf{Tanks\&Temples}\cite{knapitsch2017tanks} is a benchmark for 3D reconstruction on real-world objects of varying scales. Objects were captured from human-like perspectives under challenging illumination, shadows, and reflections. We evaluate five scenes (Barn, Caterpillar, Family, Ignatius, Truck) using dataset splits from\cite{chen2022tensorf}, with 247 training images (87\%) and 35 test images (12\%) per split.

\textbf{12Scenes}~\cite{valentin2016learning} provides RGB-D imagery from 12 rooms across four scenes, captured with depth sensors and iPad cameras. Following standard protocols, the first sequence is used for evaluation, and the others for training.

\subsection{Experimental Analysis}
\textbf{Comparison with Single RGB-Based Methods:}~\cref{tab:mipnerf360} and~\cref{tab:tanks} present quantitative comparisons across Mip-NeRF 360° and Tanks $\&$ Temples benchmarks, demonstrating SplatPose's superior accuracy in all environments. 
For Mip-NeRF 360° evaluations, our framework with only DARS-Net obtains mean angular errors of 11.1° and positional errors of 0.012, surpassing 6DGS's metrics (24.3°/0.022). With the full pipeline (DARS-Net + Pose Refinement), the errors are further reduced to 1.06° and 0.007. 
Similarly, on the Tanks $\&$ Temples dataset, using only DARS-Net achieves an angular error of 5.36° and a translation error of 0.257, surpassing 6DGS (21.7°/0.268). The full pipeline reduces these errors to 2.97° and 0.211. The full pipeline of our method achieves the best pose estimation performance on both benchmark datasets.

~\cref{fig:result_compare} presents qualitative results across various scenes, highlighting the effectiveness of our method. 
Our approach consistently produces results closely aligned with the ground truth (GT), while 6DGS demonstrates significant camera pose deviation, particularly in cluttered or complex scenes. 
These findings corroborate the quantitative results, highlighting the robustness and adaptability of our method across varying scene complexities and environments.

\textbf{Comparison with Depth- and Multi-View-Based Methods:} As shown in~\cref{tab:12scenes}, SplatPose achieves angular accuracy comparable to SplatLoc using only a single RGB image, whereas SplatLoc requires multiple views and depth information.
~\cref{tab:time} compares memory usage, training time, and inference time for different methods on the Manolis scene. 
Our method achieves a memory footprint of 264 MB, markedly lower than SplatLoc's 737 MB, by eliminating the need to store consecutive frames and leveraging a compact 3D Gaussian map representation. 
While our approach and SplatLoc exhibit comparable training times (45 minutes vs. 25 minutes) due to similar Gaussian optimization processes, both significantly outperform SCRNet's 2-day training duration, as neither relies on complex network architectures. 
Furthermore, our inference time of 6 minutes and 20 seconds surpasses SplatLoc's 9-minute runtime by bypassing initial pose retrieval. 
Although SCRNet achieves faster inference (1 minute) and lower memory usage(165 MB), this advantage is offset by its intensive computational training requirements and limited scene generalization, stemming from its data-dependent regression paradigm.

\vspace{1cm}
\begin{table*}[t]\normalsize
    \caption{The Impact of Different Stages in SplatPose on Pose Estimation Performance. We report the mean angular and translation errors (degree, u) on Mip-NeRF 360° dataset, WHERE 1u is THE object’s largest dimension. \textbf{A: DARS-NET. B:Pose Refinement.} \textbf{Bold: Best in col.}}
    \label{tab:ab}
    \centering
    \renewcommand\arraystretch{1.2}
    \linespread{1.2}\selectfont
    \resizebox{1.0\textwidth}{!}{
        \begin{tabular}{r|ccc|cccccccc}
            \toprule
            \textbf{Method} & \textbf{Baseline} & \textbf{A} & \textbf{B} & \textbf{Avg~$\downarrow$} & \textbf{Bicycle} & \textbf{Bonsai} & \textbf{Counter} & \textbf{Garden} & \textbf{Kitchen} & \textbf{Room} & \textbf{Stump}\\
            \midrule\midrule
            Exp1 & \cmark & \xmark & \xmark & $24.3 / 0.022$ & $12.1 / 0.010$ & $10.5 / 0.038$ & $19.6 / 0.043$ & $37.8 / 0.015$ & $23.2 / 0.018$ & $38.3 / 0.019$ & $28.3 / 0.009$ \\
            Exp2 & \cmark & \cmark & \xmark & $11.1 / 0.012$ & $9.14 / 0.010$ & $5.79 / 0.020$ & $9.65 / 0.022$ & $21.9 / 0.008$ & $7.91 / 0.009$ & $8.79 / 0.013$ & $14.3 / 0.005$\\
            Exp3 & \cmark & \xmark & \cmark & $10.1 / 0.009$ & $3.58 / 0.004$ & $1.10 / 0.008$ & $3.67 / 0.018$ & $25.54 / 0.007$ & $4.24 / 0.009$ & $30.9 / 0.014$ & $1.45 / \textbf{0.002}$\\
            Exp4 & \cmark & \cmark & \cmark & $\textbf{1.06 / 0.007}$ & $\textbf{0.17 / 0.003}$ & $\textbf{0.73 / 0.006}$ & $\textbf{0.52 / 0.015}$ & $\textbf{2.55 / 0.005}$ & $\textbf{0.47 / 0.008}$ & $\textbf{2.44 / 0.010}$ & $\textbf{0.53 / 0.002}$ \\
            \bottomrule
        \end{tabular}
    }
\end{table*}

\subsection{Ablation Study}
As shown in~\cref{tab:ab}, we analyze the impact of each stage of SplatPose on pose estimation performance. The baseline method (Exp1), which lacks both DARS-Net’s scoring mechanism and Pose Refinement, shows the highest average errors (24.3°, 0.022), performing poorly in sequences like Garden and Room (37.8° and 38.3° angular errors, respectively).

Exp2 introduces DARS-Net (A), reducing the average angular error by 54.4$\%$ (to 11.1°) and the translation error by 45.5$\%$ (to 0.012) compared to the baseline. Notably, in the most challenging sequence, Room, the rotation error decreases by 77.0$\%$ (from 38.3° to 8.79°). 
Exp3 replaces DARS-Net with Pose Refinement (B), achieving comparable results, with improved performance in indoor sequences like Counter (angular error reduced to 3.67°) but higher error in Garden (25.54°).

Exp4, combining DARS-Net and Pose Refinement, achieves the best performance across all sequences, reducing the average angular error by 95.6$\%$ (to 1.06°) and the translation error by 68.2$\%$ (to 0.007). In Garden, angular error drops by 93.3$\%$ (from 37.8° to 2.55°), while simpler sequences like Bonsai achieve near-perfect results (0.73°, 0.006). These results demonstrate the complementary strengths of DARS-Net and Pose Refinement, enabling robust and precise 6-DoF pose estimation.

\section{CONCLUSIONS}
This work introduces SplatPose, an advanced 6-DoF pose estimation system that builds on 3DGS with a Dual-Attention Ray Scoring Network (DARS-Net) and a coarse-to-fine pose estimation pipeline. By leveraging DARS-Net, our approach decouples positional and angular alignment in the geometric domain, effectively addressing rotational ambiguity and achieving state-of-the-art accuracy in single-image RGB pose estimation. Experiments on three public datasets—Mip-NeRF 360°, Tanks$\&$Temples, and 12Scenes—demonstrate SplatPose’s superiority over existing single RGB-based methods and depth- or multi-view-based approaches. On Mip-NeRF 360°, SplatPose achieves 10–20× lower angular errors and 3× lower translation errors than 6DGS. On Tanks$\&$Temples, it attains angular and translation errors of 2.97° and 0.211, outperforming prior methods in real-world conditions. On 12Scenes, SplatPose matches the accuracy of depth-dependent methods like SplatLoc while avoiding reliance on large image databases or depth data. Additionally, it reduces memory usage by over 64\% compared to SplatLoc and offers faster inference, enhancing practicality.







\bibliographystyle{IEEEtran}
\bibliography{ref}

\end{document}